\def\year{2022}\relax
\documentclass[letterpaper]{article} 
\usepackage{aaai22}  
\usepackage{times}  
\usepackage{helvet}  
\usepackage{courier}  
\usepackage[hyphens]{url}  
\usepackage{graphicx} 
\usepackage{natbib}  
\usepackage{caption} 
\DeclareCaptionStyle{ruled}{labelfont=normalfont,labelsep=colon,strut=off} 
\usepackage{algorithm}
\usepackage{algorithmic}

\usepackage{amssymb} 
\usepackage{amsmath} 
\usepackage{multirow} 
\usepackage{booktabs} 

\usepackage{newfloat}
\usepackage{listings}
\floatstyle{ruled}
\newfloat{listing}{tb}{lst}{}
\floatname{listing}{Listing}

\title{Learning Aligned Cross-Modal Representation \\for Generalized Zero-Shot Classification}
\author{
    Zhiyu Fang,
    Xiaobin Zhu\thanks{Correspondence author},
    Chun Yang,
    Zheng Han,
    Jingyan Qin,
    Xu-Cheng Yin
}
\affiliations{
    School of Computer \& Communication Engineering, University of Science and Technology Beijing, Beijing, China\\
    \{mr.fangzy, han970421, qinjingyanking\}@foxmail.com, \{zhuxiaobin, chunyang, xuchengyin\}@ustb.edu.cn
}

\usepackage{bibentry}

\begin{document}

\maketitle

\begin{abstract}
Learning a common latent embedding by aligning the latent spaces of cross-modal autoencoders is an effective strategy for Generalized Zero-Shot Classification (GZSC). However, due to the lack of fine-grained instance-wise annotations, it still easily suffer from the domain shift problem for the discrepancy between the visual representation of diversified images and the semantic representation of fixed attributes. In this paper, we propose an innovative autoencoder network by learning Aligned Cross-Modal Representations (dubbed ACMR) for GZSC. Specifically, we propose a novel Vision-Semantic Alignment (VSA) method to strengthen the alignment of cross-modal latent features on the latent subspaces guided by a learned classifier. In addition, we propose a novel Information Enhancement Module (IEM) to reduce the possibility of latent variables collapse meanwhile encouraging the discriminative ability of latent variables. Extensive experiments on publicly available datasets demonstrate the state-of-the-art performance of our method.
\end{abstract}

\section{Introduction}
Zero-Shot learning (ZSL) \cite{DBLP:aaai/LarochelleEB08,IEEE:RahmanSKSPF18,DBLP:aaai/WangPVFZCRC18,DBLP:ijcai/WeiDY20} aims to recognize entirely new classes without additional data labeling, which is an important issue for learning and recognition in open environments \cite{DBLP:journals/tip/HouZLSWWY20,DBLP:journals/pr/ZhuLLS21,DBLP:ZhuICCV2021}. Conventional ZSL methods try to establish the connection between the visual and semantic space by seen classes, and only the images in unseen classes are available in the testing phase. As an extension, Generalized Zero-Shot Learning (GZSL) removes the constraint in conventional ZSL by allowing images from either seen or unseen classes in the testing phase, which is a more reasonable yet challenging for real-world recognition.

\begin{figure}[t]
\centering
\includegraphics[scale=0.62]{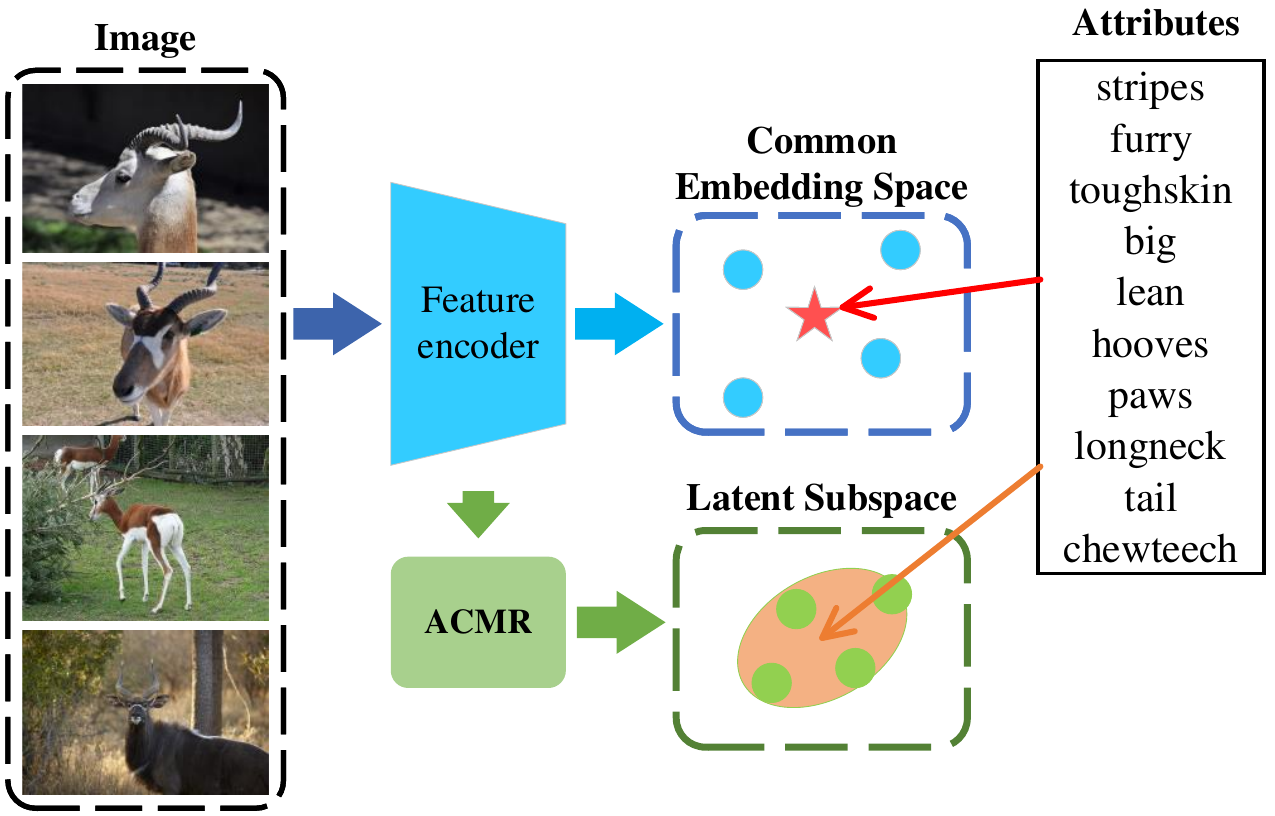}
\caption{Illustration of the overall concept of our method. Our ACMR strengthens the alignment of cross-modal latent features on latent subspaces guided by a learned classifier.}
\label{fig:overall}
\end{figure}

Recently, GZSL has received ever-increasing attention in research and industrial communities. Many GZSL methods \cite{DBLP:cvpr/ChenZ00C18,DBLP:journals/JiaZWST20,DBLP:cvpr/HanFY20,DBLP:cvpr/HuynhE20} try to learn a mapping between visual features of images and their class embeddings for knowledge transfer. Huynh and Elhamifar \cite{DBLP:cvpr/HuynhE20} adopted an attention mechanism to align attribute embeddings with local visual features. Han \textit{et al.} mapped the class-level semantic features into a redundancy-free feature space constructed from visual features. Some other methods \cite{DBLP:nips/YuL19,DBLP:eccv/VyasVP20,DBLP:journals/GaoHQCLZZS20,DBLP:cvpr/YueWS0Z21} try to generate synthetic training samples of unseen classes for GZSL. Vyas \textit{et al.} \cite{DBLP:eccv/VyasVP20} proposed a semantic regularized loss for conditional WGAN to generate visual features for bridging the distribution gap between seen and unseen classes. However, because the distribution of visual features in seen and unseen classes are generally distinguished, unidirectional mappings can easily induce the domain shift problem \cite{DBLP:journals/FuHXG15,DBLP:cvpr/ZhuWS19}.

To alleviate the domain shift problem, some methods \cite{DBLP:cvpr/XianA0N0S16,DBLP:iccv/TsaiHS17,DBLP:cvpr/VermaAMR18} try to learn common latent spaces by cross-modal alignment restriction. In this paradigm, Variational AutoEncoder (VAE) based architectures are popular for the capability of fitting data distribution in latent space. Schonfeld \textit{et al.} \cite{DBLP:cvpr/SchonfeldESDA19} employed two VAEs to learn the joint reconstruction between visual features and semantic features in the common latent space. Ma \textit{et al.} \cite{DBLP:aaai/MaH20} incorporated a deep embedding network and modified variational autoencoder to learn a latent space shared by both image features and class embeddings. Li \textit{et al.} \cite{DBLP:aaai/LiXWD21} respectively decoupled category-distilling factors and category-dispersing factors from visual and semantic feature for GZSL. However, due to the lack of fine-grained instance-wise annotations, the aligned relationships across different modalities tend to be confused on the common embedding space without structural restriction, which is detrimental for generating effective latent features to train a classifier. A plausible solution is to further restrict cross-modality alignment on subspace, especially for image classification.

From the above-mentioned observations, we propose an innovative autoencoder network by learning Aligned Cross- Modal Representations (dubbed ACMR) for GZSC. Figure \ref{fig:overall} illustrates the motivation of our method. Our ACMR is constructed on the popular Variational Auto-Encoders (VAEs) to encode and decode features from different modalities and align them on the latent space. To be specific, a novel Vision-Semantic Alignment (VSA) is proposed to further strengthen the alignment of cross-modal latent features on latent subspace guided by a learned classifier. In addition, a novel Information Enhancement Module (IEM) is proposed to reduce the possibility of latent variables collapse by maximizing joint distribution, meanwhile encouraging the discriminative ability of latent variables. Finally, the aligned cross-modal representations are used to train a softmax classifier for GZSC.

In summary, our main contributions are three-fold:
\begin{itemize}
\item We propose an innovative autoencoder network by learning Aligned Cross-Modal Representations for GZSC. Experimental results verify the state-of-the-art performance of our method on four publicly available datasets.
\item Proposing an innovative Vision-Semantic Alignment (VSA) method to strengthen the alignment of cross-modal latent features on latent subspace guided by a learned classifier.
\item Proposing a novel Information Enhancement Module (IEM) to reduce the possibility of latent variables collapse meanwhile enhancing the discriminative ability of latent variables.
\end{itemize}

\section{Related Work}
\subsection{Generalized Zero-Shot Learning}
GZSL is a more realistic than ZSL because it not only learns information which can be adapted to unseen classes but also apply to the testing data from seen classes. The classical GZSL methods learn a mapping function and can be roughly divided into three categories.

One type of methods directly extracts visual features for mapping to semantic space. DAP and IAP \cite{datasetAWA} establish the rigid correspondence between original visual features and attribute vectors for GZSL. The popular attention mechanisms \cite{DBLP:journals/isci/ZhuLLLD20,DBLP:ZhuTITS} also have been adopted to distinguish informative local visual features for GZSL, such as \cite{DBLP:cvpr/Xie0J0ZQY019,IEEE:NiuLCJVAZL19,DBLP:cvpr/HuynhE20,DBLP:cvpr/Liu00H00H21,DBLP:aaai/GeXM021}. Considering the distribution consistency of semantic features in all categories, some methods aims to extract semantic features from textual descriptions for mapping to visual space. DeViSE \cite{DBLP:nips/FromeCSBDRM13} maps the visual features into word2vector embedding space learned by Skip-gram Language Model \cite{DBLP:iclr/abs-1301-3781}. Moreover, the generative models \cite{DBLP:eccv/FelixKRC18,DBLP:nips/WanCLYZY019,DBLP:eccv/VyasVP20,IEEE:nnls/FengZ21,DBLP:cvpr/YueWS0Z21} try to generate synthetic images of unseen classes for GZSL.

Due to the above methods often conduct mapping in a unidirectional manner, which tend to easily cause the domain shift problem. Therefore, the common-space based methods try to map different features into common spaces for connecting visual features with semantic features \cite{DBLP:cvpr/AkataRWLS15,DBLP:journals/AkataPHS16,DBLP:cvpr/SungYZXTH18,DBLP:cvpr/Keshari0V20}. CVAE \cite{DBLP:cvpr/MishraRMM18} utilizes conditional VAE to encode the concatenation feature of visual and semantic embedding vectors. OCD \cite{DBLP:cvpr/Keshari0V20} designs an over-complete distribution of seen and unseen classes to enhance class separability. Although these methods extract the effective features on the common space, they often neglect explicitly aligning different modalities.

\subsection{Cross-Modal Alignment Models for GZSL}
Recent cross-alignment based methods of GZSL generally focus on establishing a shared latent space for visual and semantic features via generative adversarial network (GAN) \cite{DBLP:nips/GoodfellowPMXWOCB14} or variational auto-encoder (VAE) \cite{ArXiv:kingma2013auto}. The pioneer GAN-based method, SJE \cite{DBLP:cvpr/AkataRWLS15} learns a convolutional GAN framework to synthesize images from high-capacity text via minimax compatibility between image and text. LiGAN \cite{DBLP:cvpr/LiJLD0H19} employs GAN to map features into synthetic sample space. f-VAEGAN-D2 \cite{DBLP:cvpr/XianSSA19} learns the marginal feature distribution of unlabeled images via a conditional generative model and an unconditional discriminator. cycle-GAN \cite{DBLP:eccv/FelixKRC18} and EPGN \cite{DBLP:cvpr/YuJHZ20} strengthen the alignment of latent features in their respective modalities by cross-reconstruction. However, GAN-based models are difficult to optimize parameters during the training process.

\begin{figure*}[t]
\centering
\includegraphics[scale=0.8]{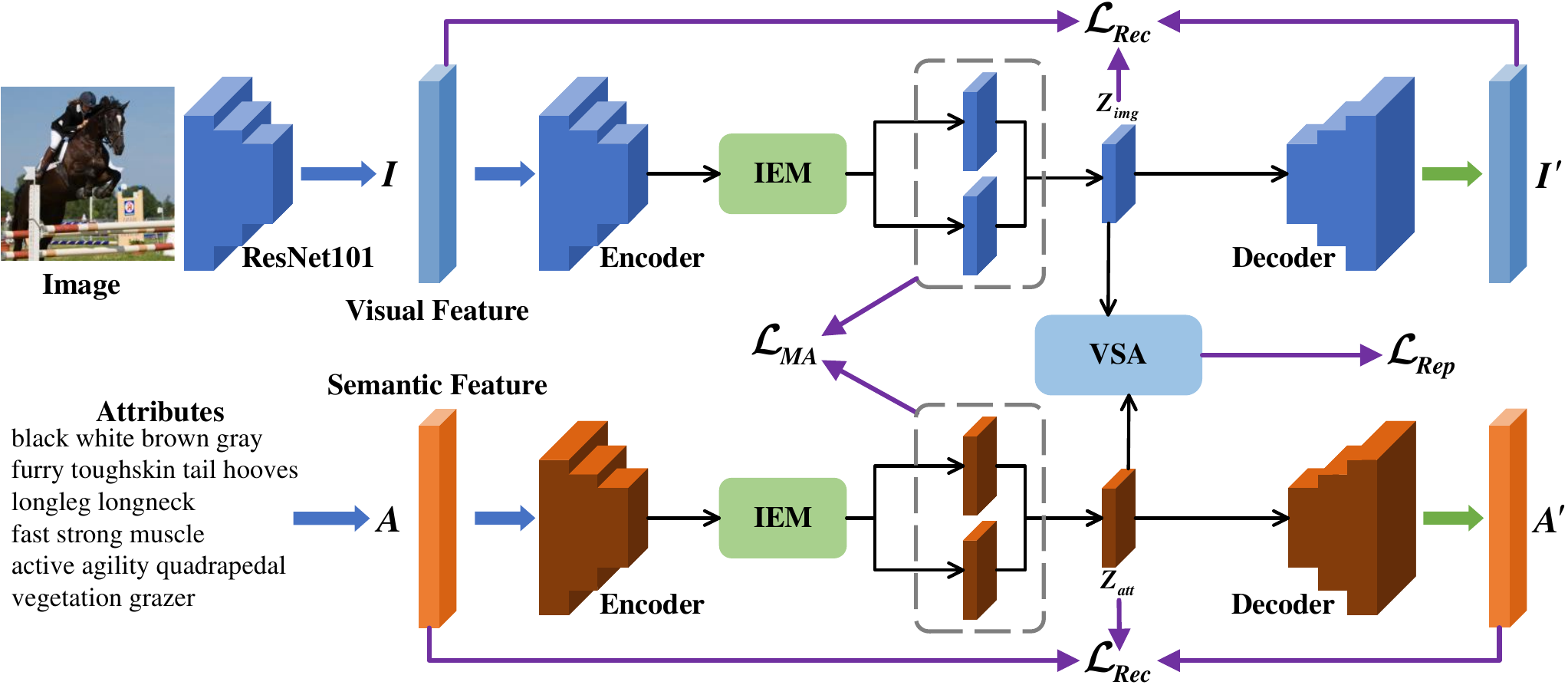}
\caption{Architecture of our Aligned Cross-Modal Representations (ACMR) network. Two parallel VAEs as backbones extract visual and semantic latent representations, respectively. IEM (detailed in Figure \ref{fig:iem}) aims to learn more discriminative latent representation. VSA (detailed in Figure \ref{fig:vsa}) restrains the alignment between two modalities by classification optimization.}
\label{fig:network}
\end{figure*}

VAE is an autoencoder for learning the conditional probability distribution from training data, which helps to generate samples having certain desired properties. CADA-VAE \cite{DBLP:cvpr/SchonfeldESDA19} combines cross-modal and cross-reconstruction, and leverages two independent VAEs to align features via minimizing distance between the latent distributions of visual and semantic features. Based on CADA-VAE, Li \textit{et al.} \cite{DBLP:mm/LiJ0DL020} learns modality-invariant latent representations by maximizing mutual information and entropy on latent space. DE-VAE \cite{DBLP:aaai/MaH20} adopts a deep embedding model to learn the mapping from the semantic space to the visual feature space. Disentangled-VAE \cite{DBLP:aaai/LiXWD21} decouples the latent features by shuffling classification-irrelevant information to obtain discriminative representations. Although the VAE-based methods achieve promising performance, the aligned relationships across different modalities tend to be confused on the common embedding space without structural restriction since the lack of fine-grained instance-wise annotations.

\section{Methodology}
In this section, we first present the definition of GZSL and the analysis of cross-modal alignment in common space. Then we briefly introduce VAE as the basic building block of our model, and elaborate our method. Finally, we summarize the loss function and training process of our method.

\subsection{Generalized Zero-Shot Learning}
Suppose that $X$$=$$\{X^{S},\ X^{U}\}$ denotes the visual space of images, $A$$=$$\{A^{S},\ A^{U}\}$ denotes the semantic space of attributes, and $Y$$=$$\{Y^{S},\ Y^{U}\}$ denotes the corresponding label set. Notably, $Y^{S}$ and $Y^{U}$ are disjoint, i.e., $Y\cap Y^{U}$$=$$\emptyset$. Consequently, we can individually collect training dataset $D_{s}$$=$$\{x_{i}^{s}, a_{i}^{s}, y_{i}^{s}\}_{i=1}^{N}$ and testing dataset $D_{u}$$=$$\{x_{i}^{u}, a_{i}^{u}, y_{i}^{u}\}_{i=1}^{M}$, where $x_{i}^{s}, x_{i}^{u} \in X$ is the $i$-th visual feature, $a_{i}^{s}, a_{i}^{u} \in A$ is the $i$-th semantic feature, and $y_{i}^{s}, y_{i}^{u} \in Y$ is their corresponding labels, of seen/unseen classes, respectively. The task of ZSL aims to learn a classifier $f_{Z S L}: X \rightarrow Y^{U}$ for recognizing a testing instance $x$ of unseen classes. For adapting to both seen and unseen classes, GZSL adopts a more realistic setting and learn a classifier $f_{G Z S L}: X \rightarrow Y^{U} \cup Y^{S}$.

\subsection{Cross-Modal Alignment}
Let $Z_{x}, Z_{a}$ respectively denote the feature vectors of visual and semantic modalities, the joint distribution $P(Z_{x}, Z_{a})$ can assess the degree of cross-modal alignment. For example, if two modalities are completely aligned $P(Z_{x}, Z_{a})=$ 1 , otherwise $0 \leq P(Z_{x}, Z_{a})<1$. Optimizing cross-modal alignment can be mathematically formulated as:
\begin{equation}
\begin{aligned}
&2\log p(z_x,z_a)\ge\log p(z_x|z_a)+\log p(z_a|z_x)+\mathcal{H}(X)\\
&+\mathbb{E}_{p(x)}[\log p(z_x,X)]+\mathbb{E}_{p(a)}[\log p(z_a,A)]+\mathcal{H}(A)
\label{eq:2logp}
\end{aligned}
\end{equation}
where $\mathbb{E}(\cdot)$ represents expectation, $\mathcal{H}(\cdot)$ represents entropy.

\subsection{Aligned Cross-Modal Representation Network}
For conducting cross-modal alignment, we can train a deep learning model for mapping image features and attributes into a common space. According to the optimization objective (Equation \ref{eq:2logp}), the common space can be utilized to align the features by the first two items, the features on the common space can contain discriminative information corresponding inputs by the middle two items, and the features can be utilized to reconstruct original modalities by the last two items. Consequently, we summarize Equation \ref{eq:2logp} into three parts. The first two log-likelihood items $\log p(\cdot)$ can be realized by a mutual alignment constraint $C_{M A}$, the middle two expectation items $\mathbb{E}$ can be realized by a representation constraint $C_{R e p}$, and the last two entropy items $\mathcal{H}$ can be realized by a reconstruction constraint $C_{R e c}$. Hence, the total optimization objective can be simplified as:
\begin{equation}
\mathcal{O}=C_{MA}+C_{Rep}+C_{Rec}
\label{eq:o}
\end{equation}

\textbf{Variational Auto-Encoder.} Since the excellent ability of fitting data distribution, VAE is widely used in cross modal alignment. We employ two independent VAEs (shown in Figure \ref{fig:network}) composed of multilayer perceptron (MLP) to respectively establish visual and semantic modality. In VAE, the encoders are used to generate the latent variables of each modality, and the decoders are used to reconstruct the latent vectors. Therefore, we can estimate the reconstruction constraint $C_{R e c}$ with VAE loss, which can be formulated as:
\begin{equation}
\begin{aligned}
\mathcal{L}_{Rec}&= \mathbb{E}_{q_{\phi}(z_x|x)}[\log p(x|z_x)]-\alpha D_{KL}(q_{\phi}(z_x|x) \| p_{\theta}(z_x)) \\
&+\mathbb{E}_{q_{\phi}(z_a|a)}[\log p(a|z_a)]-\alpha D_{KL}(q_{\phi}(z_a|a) \| p_{\theta}(z_a))
\label{loss:rec}
\end{aligned}
\end{equation}
where $q_{\phi}(z_{x}|x)$ denotes latent feature distributions of visual modality, $q_{\phi}(z_{a}|a)$ denotes latent feature distributions of semantic modality, $p(x|z_{x})$ and $p(a|z_{a})$ denote the observation distributions of corresponding modalities, $\alpha$ weights the KL Divergence. The priors $p_{\theta}(z_{x})$ and $p_{\theta}(z_{a})$ are both standard Gaussian distributions.

Because VAE models the distribution of hidden variables, we can transform the mutual alignment constraint $C_{M A}$ of maximizing log likelihood into minimizing the Wasserstein distance between the latent Gaussian distributions of semantic modality and visual modality. Hence, the mutual alignment constraint can be formulated as:
\begin{equation}
\mathcal{L}_{MA}=\left(\|\mu_x-\mu_a\|_2^2+\left\|\Sigma_x^{\frac{1}{2}}-\Sigma_a^{\frac{1}{2}}\right\|_{\text {Frobenius }}^2\right)^{\frac{1}{2}}
\label{loss:ma}
\end{equation}
where $\mu_{x}$ and $\Sigma_{x}$ are all predicted by visual encoder, while $\mu_{a}$ and $\Sigma_{a}$ are all predicted by semantic encoder.

\begin{figure}[t]
\centering
\includegraphics[scale=0.4]{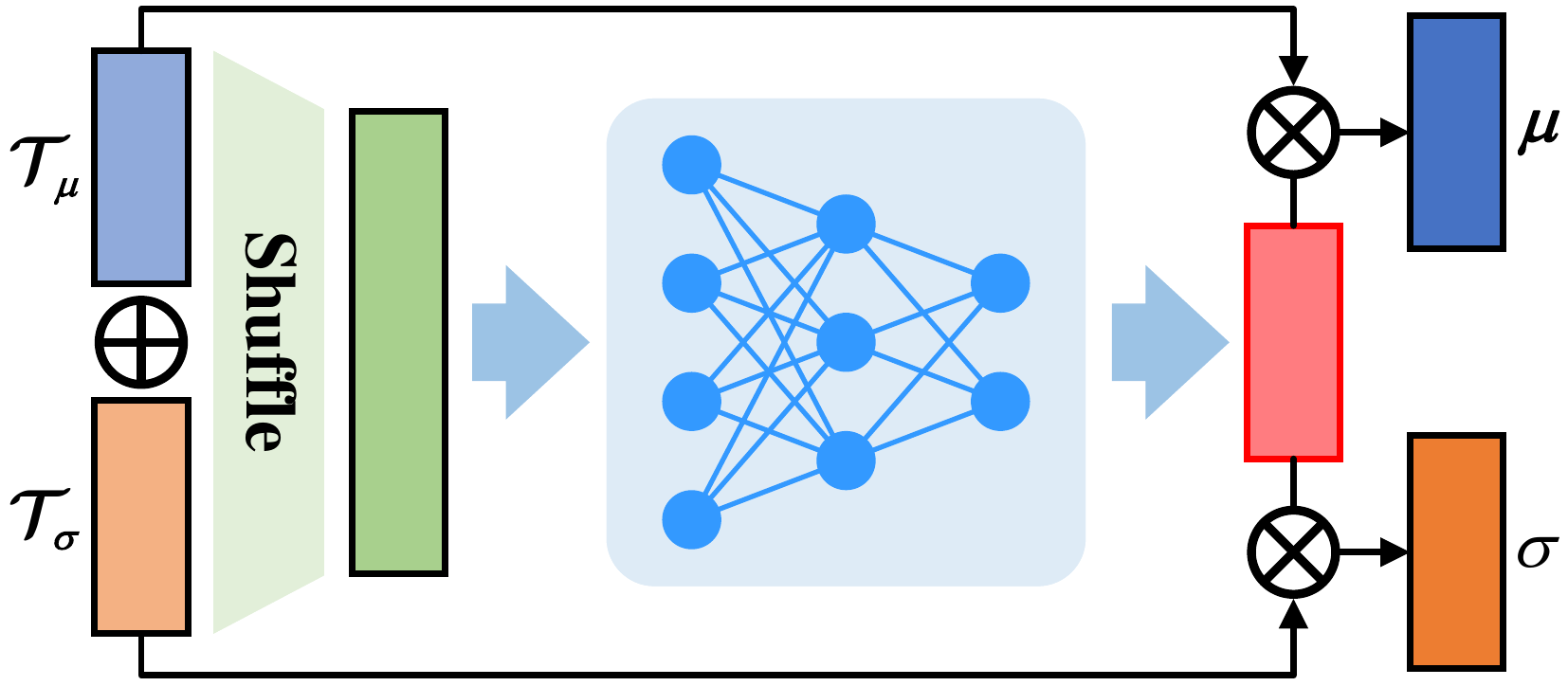}
\caption{Architecture of our IEM.}
\label{fig:iem}
\end{figure}

\textbf{Information Enhancement Module.} Because of the ``latent variable collapse'' problem \cite{DBLP:icml/AlemiPFDS018}, the distribution of latent vector $p(Z|I)$ generated by VAE can easily degrade into a standard normal distribution $\mathcal{N}(0,1)$, which is detrimental to learn discriminative representations of latent variables for classification. From this key observation, we propose a novel Information Enhancement Module (IEM, shown in Figure \ref{fig:iem}) to magnify the joint probability between observation $I$ and the inferred latent variables $Z$.

According to \cite{DBLP:cvpr/ChenBZM19}, we randomly shuffle the features of encoder to extract discriminative features. For obtaining the joint probability, we develop a MLP with one hidden layer to learn the marginal probability of $I$, and multiply it with the conditional probability of $Z$. Mathematically, our IEM can be formulated as:
\begin{equation}
\begin{aligned}
p(z_x,X)&=p(E(X;\varphi_x)\mid X)p(\varPsi(X;\psi_x)) \\
p(z_a,A)&=p(E(A;\varphi_a)\mid A)p(\varPsi(A;\psi_a))
\label{eq:jiontp}
\end{aligned}
\end{equation}
where $p(z_x,X)$, $p(z_a,A)$ denote latent variable distributions with enhanced mutual information of visual modality and semantic modality; $E(X;\varphi_x), E(A;\varphi_a)$ respectively denote visual modality encoder and semantic modality encoder; $\varPsi(X; \psi_x), \varPsi(A; \psi_a)$ are marginal probability perceptrons of corresponding modalities.

When optimizing the vanilla VAE with joint probability, KL divergence can be re-defined as:
\begin{equation}
\begin{aligned}
D_{KL}&=\int q_{\phi}(z,i) \log\frac{q_{\phi}(z,i)}{p_{\theta}(z)}dz \\
&=q_{\phi}(i)D_{KL}(q_{\phi}(z|i) \| p_{\theta}(z))+q_{\phi}(i) \log q_{\phi}(i)
\label{eq:dkl}
\end{aligned}
\end{equation}
According to Equation \ref{loss:rec}, the latent variable will collapse with the decrease of $\mathcal{L}_{Rec}$. Then, $D_{KL}(q_{\phi}(z|i) \| p_{\theta}(z))$ will be approximate to 0. But with $q_{\phi}(i) \log q_{\phi}(i)$ containing distribution of observation $i$ in re-defined $D_{KL}$, $z$ will not degrade to $\mathcal{N}(0,1)$ and tend to contain more effective information about true data distribution.

\begin{figure*}[t]
\centering
\includegraphics[scale=0.5]{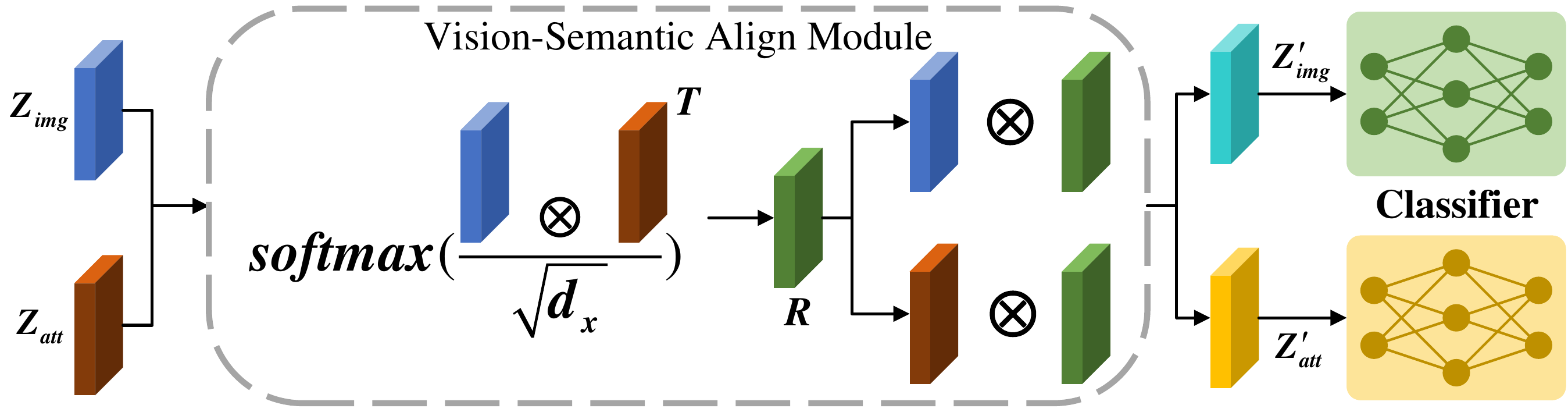}
\caption{Architecture of Vision-Semantic Alignment Module (VSA). $d_x$ denotes normalization factor, $T$ denotes transpose operator, and $R$ denotes joint weighting coefficient of latent variables.}
\label{fig:vsa}
\end{figure*}

\textbf{Vision-Semantic Alignment.} Vision-Semantic Alignment (VSA, shown in Figure \ref{fig:vsa}) is proposed to align the multi-modal latent variables $Z$ after information enhancement. Since the lack of fine-grained instance-wise annotation, images of each class share a pre-class attribute vector, the alignment between different modalities tend to be confused on the latent space without any structural restriction. To tackle this issue, we further constrain the latent features on latent subspace guided by a learned classifier to strengthen the alignment. We introduce classification information $Y$ into the optimization function and maximize the joint distribution $p(Z, I, Y)$. Then the optimization function can be obtained by replacing $p(z_x, X), p(z_a, A)$ in the middle two items of Equation \ref{eq:2logp} with $p(z_x, X, Y)$, $p(z_a, A, Y)$. Then the two expectation items of Equation \ref{eq:2logp} can be respectively formulated as:
\begin{equation}
\begin{aligned}
\mathbb{E}&_{p(x)}[\log p(z_x, X)]= \\
&\mathbb{E}_{p(y)}\mathbb{E}_{p(x)}[\log p(Y|z_x,X)]+\mathbb{E}_{p(y)}\log p(z_x,X) \\
\mathbb{E}&_{p(a)}[\log p(z_a, A)]= \\
&\mathbb{E}_{p(y)}\mathbb{E}_{p(a)}[\log p(Y|z_a,A)]+\mathbb{E}_{p(y)}\log p(z_a,A)
\label{eq:elog}
\end{aligned}
\end{equation}
where $p(Y|z_{x}, X), p(Y|z_{a}, A)$ denote classification probability of corresponding modalities in latent space, the last $\mathbb{E}(\cdot)$ represent joint expectation. This formula highlights that maximizing the optimization objective is equivalent to maximize the conditional probability $p(Y|z_{x}, X)$ and $p(Y|z_{a}, A)$ and to maximize the expectation of the joint distribution $\mathbb{E}(\cdot) .$ Hence, $\mathbb{E}_{p(y)}\mathbb{E}_{p(x)}[\cdot]$ item can be approximated by a classification loss based on latent variables, $\mathbb{E}(\cdot)$ item can be approximated by a reconstruction loss based on latent variables after IEM. In this way, we can constrain the generation of latent variables by the classifier to improve its discriminability.

By replacing the expectation items in representation constraint $C_{Rep}$ with latent variable representation after information enhancement (shown in Equation \ref{eq:elog}). The new $C_{Rep}$ are implemented by IEMs and VSA modules, and the entropy items as reconstruction are merged into VAE. Hence we can denote the representation loss as:
\begin{equation}
\begin{aligned}
\mathcal{L}_{Rep}=&\frac{1}{N}\sum_{i=1}^{N}L_{cls}(y_i^s, f(z_i^x; W_x))+\\
&\frac{1}{N}\sum_{i=1}^{N}L_{cls}(y_i^s, f(z_i^a; W_a))
\label{loss:rep}
\end{aligned}
\end{equation}
where $z_i^x$, $z_i^a$, $y_i^s$ are visual latent variable, semantic variable, and label in training set, $L_{cls}$ is the classification loss, $f$ is the classifier composed of two independent block.

\textbf{Loss function.} The overall loss of our method can be formulated as:
\begin{equation}
\mathcal{L}=\mathcal{L}_{Rec}+\beta\mathcal{L}_{MA}+\lambda\mathcal{L}_{Rep}
\label{loss:all}
\end{equation}
where $\mathcal{L}_{Rec}$ is the reconstruction loss of intra-modalities, $\mathcal{L}_{MA}$ is the mutual alignment loss of inter-modalities, and $\mathcal{L}_{Rep}$ is the classification loss of latent representations. $\beta$ and $\lambda$ are two hyper-parameters training by an annealing schedule \cite{DBLP:conll/BowmanVVDJB16}.

\section{Experiments}

\begin{table}[t]
\begin{center}
\caption{Common datasets in GZSL, in terms of dimensionality of visual features $\mathcal{D}_{I}$, dimensionality of semantic features $\mathcal{D}_{A}$, number of all instances $\mathcal{N}_{X}$, number of seen classes $\mathcal{N}_{S}$ and number of unseen classes $\mathcal{N}_{U}$.}
\label{tab:dataset}
\scalebox{1}{
\begin{tabular}{*{5}c}
\toprule
Dataset & $\mathcal{D}_{I}$ & $\mathcal{D}_{A}$ & $\mathcal{N}_{X}$ &$\mathcal{N}_{S}/\mathcal{N}_{U}$ \\
\midrule
CUB & \multirow{4}*{2048} & 312 & 11788 & 150/50 \\
SUN & & 102 & 14340 & 707/10 \\
AwA1 & & 85 & 30475 & 40/10 \\
AwA2 & & 85 & 37322 & 40/10 \\
\bottomrule
\end{tabular}}
\end{center}
\end{table}

\subsection{Datasets and Evaluation Protocol}
We conduct experiments on four benchmark datasets: CUB \cite{datasetCUB}, SUN \cite{datasetSUN}, AwA1 \cite{datasetAWA}, and AwA2 \cite{datasetAWA2}. The detailed information of the datasets is summarized in Table \ref{tab:dataset}. Specifically, CUB is a fine-grained categorized dataset collected from professional bird websites, which covers 200 categories and each category has a 312-dimensional attribute vector. SUN is a fine-grained scene understanding dataset, which is widely used in objective detection, image classification, and so on. In SUN, images are annotated with 102-dimensional attribute. AwA1 is an animal image dataset that uses 85-dimensional attribute to describe 50 categories. AwA2 is a more sophisticated variant of AwA1, which collects 37,322 images from public web sources. AwA1 and AwA2 have no overlapping images.

For a fair comparison, we adopt the setting as in \cite{datasetAWA2} for training and testing. The evaluation metrics in generalized datasets setting include the average classification accuracy (ACA) of the samples from seen/unseen classes, the harmonic mean of $ACA^{S}$ and $ACA^{U}$, details are listed as followings:
\begin{itemize}
\item $ACA^{U}$: the average accuracy of per-class on test images from unseen classes, which represents the capacity of classifying unseen classes samples.
\item $ACA^{S}$: the average accuracy of per-class on test images from seen classes, which is used to represent the capacity of classifying seen classes samples.
\item $H$: the harmonic mean value, which is formulated as
\begin{equation}
H=\frac{2 \times ACA^{U} \times ACA^{S}}{ACA^{U}+ACA^{S}}
\end{equation}
\end{itemize}

\subsection{Implementation Details}
Following the setting in other methods \cite{datasetAWA2,DBLP:cvpr/SchonfeldESDA19}, ResNet-101 pre-trained on ImageNet is used to extract the 2048 dimensional visual features. In IEM, the hidden layer of MLP has 99 units. We respectively utilize 1560, 1680 hidden units for visual encoder and decoder, and 1450, 665 hidden units for semantic encoder and decoder. The size of our aligned cross-modal representation in latent space is 64 for all datasets. The final softmax classifier includes a fully connected layer and a non-linear activation layer. The input size of the fully connected layer is 64, and the output size is equal to the total number of all categories. Our approach is implemented with PyTorch 1.5.0 and trained for 100 epochs by the Adam optimizer \cite{DBLP:journals/corr/KingmaB14}.We set learning rate as 1.5e-04 for training VAEs, 3.3e-05 for training IEMs, 7.4e-03 for training VSA, 0.5e-03 for training softmax classifier. For all datasets, the batch size of ACMR is set to 50 and the batch size of final softmax classifier is set to 32.

\begin{table*}[t]
\begin{center}
\caption{Performance of classification on four benchmarks under PS (proposed split). ts=ACA on $U={ACA}^U$, $S={ACA}^S$, $H$ is the harmonic mean. (The best H is marked {\bf bold}).}
\label{tab:gzsl}
\scalebox{0.87}{
\begin{tabular}{cl*{12}{c}}
\toprule
& \multicolumn{1}{c}{\multirow{2}*{Model}} & \multicolumn{3}{c}{CUB} & \multicolumn{3}{c}{SUN} & \multicolumn{3}{c}{AwA1} & \multicolumn{3}{c}{AwA2} \\
& & U & S & H & U & S & H & U & S & H & U & S & H \\
\midrule
\multirow{3}*{Embedding} & SP-AEN \cite{DBLP:cvpr/ChenZ00C18} & 34.7 & 70.6 & 46.6 & 24.9 & 38.6 & 30.3 & - & - & - & 23.3 & 90.9 & 37.1 \\
& AREN \cite{DBLP:cvpr/Xie0J0ZQY019} & 38.9 & 78.7 & 52.1 & 19.0 & 38.8 & 25.5 & - & - & - & 15.6 & 92.9 & 26.7 \\
& DAZLE \cite{DBLP:cvpr/HuynhE20} & 42 & 65.3 & 51.1 & 25.7 & 82.5 & 25.8 & - & - & - & 25.7 & 82.5 & 39.2 \\
\hline
\multirow{7}*{GAN} & cycle-CLSWGAN \cite{DBLP:eccv/FelixKRC18} & 59.3 & 47.9 & 53 & 33.8 & 47.2 & 39.4 & 63.4 & 59.6 & 59.8 & - & - & - \\
& f-CLSWGAN \cite{DBLP:cvpr/XianLSA18} & 57.7 & 43.7 & 49.7 & 36.6 & 42.6 & 39.4 & 61.4 & 57.9 & 59.6 & 53.8 & 68.2 & 60.2 \\
& LiGAN \cite{DBLP:cvpr/LiJLD0H19} & 46.5 & 57.9 & 51.6 & 42.9 & 37.8 & 40.2 & 52.6 & 76.3 & 62.3 & 54.3 & 68.5 & 60.6 \\
& f-VAEGAN-D2 \cite{DBLP:cvpr/XianSSA19} & 60.1 & 48.4 & 53.6 & 38 & 45.1 & 41.3 & 70.6 & 57.6 & 63.5 & - & - & - \\
& DASCN \cite{DBLP:nips/NiZ019} & 59 & 45.9 & 51.6 & 38.5 & 42.4 & 40.3 & 68 & 59.3 & 63.4 & - & - & - \\
& LsrGAN (Vyas et al. 2020) & 59.1 & 48.1 & 53 & 37.7 & 44.8 & 40.9 & 74.6 & 54.6 & 63 & 74.6 & 54.6 & 63 \\
& E-PGN \cite{DBLP:cvpr/YuJHZ20} & 57.2 & 48.5 & 52.5 & - & - & - & 86.3 & 52.6 & 65.3 & 83.6 & 48 & 61 \\
\hline
\multirow{6}*{VAE} & SE \cite{DBLP:cvpr/VermaAMR18} & 53.3 & 41.5 & 46.7 & 30.5 & 40.9 & 34.9 & 67.8 & 56.3 & 61.5 & 68.1 & 58.3 & 62.8 \\
& CVAE \cite{DBLP:cvpr/MishraRMM18} & – & – & 34.5 & – & – & 26.7 & – & – & 47.2 & – & – & 51.2 \\
& CADA-VAE \cite{DBLP:cvpr/SchonfeldESDA19} & 51.6 & 53.5 & 52.4 & 47.2 & 35.7 & 40.6 & 57.3 & 72.8 & 64.1 & 55.8 & 75 & 63.9 \\
& OCD \cite{DBLP:cvpr/Keshari0V20} & 59.9 & 44.8 & 51.3 & 42.9 & 44.8 & 43.8 & - & - & - & 73.4 & 59.5 & 65.7 \\
& DE-VAE \cite{DBLP:aaai/MaH20} & 52.5 & 56.3 & 54.3 & 45.9 & 36.9 & 40.9 & 59.6 & 76.1 & 66.9 & 58.8 & 78.9 & 67.4 \\
& Disentangled-VAE \cite{DBLP:aaai/LiXWD21} & 51.1 & 58.2 & 54.4 & 36.6 & 47.6 & 41.4 & 60.7 & 72.9 & 66.2 & 56.9 & 80.2 & 66.6 \\
\hline
{\bf Ours} & ACMR & 53.1 & 57.7 & {\bf55.3} & 49.1 & 39.5 & {\bf43.8} & 59.4 & 77.6 & {\bf67.5} & 60.0 & 80.2 & {\bf68.7} \\
\bottomrule
\end{tabular}}
\end{center}
\end{table*}

\subsection{Comparing with the State-of-the-Arts}
\textbf{Baseline models.} We compare our model with 15 state-of-the-art (SOTA) models. Among them, SP-AEN \cite{DBLP:cvpr/ChenZ00C18}, AREN \cite{DBLP:cvpr/Xie0J0ZQY019}, DAZLE \cite{DBLP:cvpr/HuynhE20} are deep embedding based methods, cycle-CLSWGAN \cite{DBLP:eccv/FelixKRC18}, f-CLSWGAN \cite{DBLP:cvpr/XianLSA18}, LiGAN \cite{DBLP:cvpr/LiJLD0H19}, f-VAEGAN-D2 \cite{DBLP:cvpr/XianSSA19}, DASCN \cite{DBLP:nips/NiZ019}, LsrGAN \cite{DBLP:eccv/VyasVP20}, E-PGN \cite{DBLP:cvpr/YuJHZ20} are GAN based methods, and SE \cite{DBLP:cvpr/VermaAMR18}, CVAE \cite{DBLP:cvpr/MishraRMM18}, CADA-VAE \cite{DBLP:cvpr/SchonfeldESDA19}, OCD \cite{DBLP:cvpr/Keshari0V20}, DE-VAE \cite{DBLP:aaai/MaH20}, Disentangled-VAE \cite{DBLP:aaai/LiXWD21} are VAE-based methods.

\begin{table}[t]
\begin{center}
\caption{Ablation study on different module combinations. The results are $H$ (harmonic mean) on four datasets.}
\label{tab:abla}
\scalebox{1}{
\begin{tabular}{*{5}c}
\toprule
Model & CUB & SUN & AwA1 & AwA2 \\
\midrule
Dual-VAE & 32.7 & 38.3 & 52.4 & 66.1 \\
IE-VAE & 36.1 & 39.0 & 65.1 & 66.6 \\
VSA-VAE & 54.1 & 43.5 & 65.7 & 67.9 \\
ACMR & {\bf55.2} & {\bf43.8} & {\bf67.3} & {\bf68.7} \\
\bottomrule
\end{tabular}}
\end{center}
\end{table}

The detailed results are listed in Table \ref{tab:gzsl}. In our method, the value of $H$ can reach 55.3\% on CUB, 43.8\% on SUN, 67.5\% on AwA1, and 68.7\% on AwA2, respectively. Specifically, our method respectively outperforms the original benchmark CADA-VAE by 2.9\% on CUB, by 3.2\% on SUN, by 3.4\% on AwA1, and by 4.8\% on AwA2 in terms of $H$. Compared with deep embedding based methods, our ACMR outperforms DAZLE on three datasets. Especially, our ACMR significantly increases $H$ by 18\% and 29.5\% on SUN and AwA2, respectively. Moreover, our method outperforms APN \cite{DBLP:nips/XuXWSA20} by 3.1\% (H) on AwA2, by 0.1\% (H) on SUN, and outperforms RGEN \cite{DBLP:eccv/Xie0ZZZYQ020} by 7\% (H) on SUN. Compared with GAN-based methods, our ACMR outperforms LsrGAN by 2.3\% on CUB, by 2.9\% on SUN, by 4.5\% on AwA1, and by 5.7\% on AwA2 in terms of $H$, respectively. We can respectively increase $H$ by 2.8\% on CUB, by 2.2\% on AwA1, and by 7.7\% on AwA2, compared with E-PGN whose $H$ are 52.5\%, 65.3\% and 61\% corresponding datasets. Compared with VAE-based methods, our ACMR outperforms Disentangled-VAE by 0.9\% on CUB, by 2.4\% on SUN, by 1.3\% on AwA1, and by 2.1\% on AwA2 in terms of $H$, respectively. Compared with DE-VAE, we can increase $H$ from 54.3\% to 55.3\% on CUB, from 40.9\% to 43.8\% on SUN, from 66.9\% to 67.5\% on AwA1 and from 67.4\% to 68.7\% on AwA2. Obviously, our method achieves SOTA performance and outperform other common space based methods with a great margin.

\subsection{Ablation Study}
In this section, we conduct ablation study to evaluate the effectiveness and necessity of our proposed components, including backbone, IEM, VSA, and the hyper-parameters of weighting loss.

\textbf{About the IEM and VSA.} We present the results under different combinations of our proposed modules in ACMR, including Dual-VAE (only two VAEs), IE-VAE (two VAEs and two IEMs), VSA-VAE (two VAEs and VSA module) and our original ACMR. The detailed results are listed in Table \ref{tab:abla}.

From Table \ref{tab:abla}, we can see that VSA-VAE outperforms Dual-VAE by 21.4\% on CUB, by 5.2\% on SUN, by 13.3\% on AwA1, and by 1.8\% on AwA2 in terms of $H$. It means that discriminative representations can be learned by aligning cross modal features guided by classifier. In addition, compared with Dual-VAE, IE-VAE increases $H$ from 32.7\% to 36.1\% on CUB, from 38.3\% to 39.0\% on SUN, from 52.4\% to 65.1\% on AwA1, and from 66.1\% to 66.6\% on AwA2, respectively. Compared with VAS-VAE, ACMR increases $H$ from 54.1\% to 55.2\% on CUB, from 43.5\% to 43.8\% on SUN, from 65.7\% to 67.3\% on AwA1, and from 67.9\% to 68.7\% on AwA2, respectively. It means that the information enhancement module can effectively strengthen the discriminative ability of latent variables.

\begin{figure*}[t]
\centering
\includegraphics[scale=0.29]{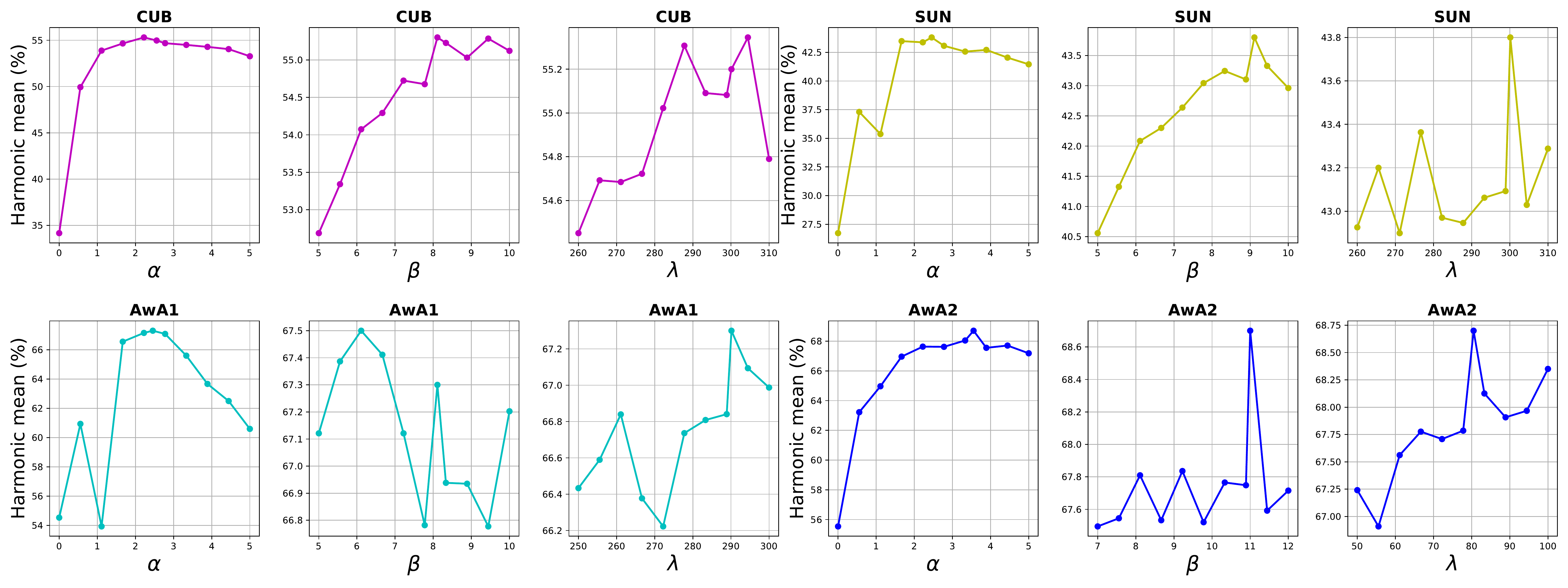}
\caption{Influence of the weighting coefficients of losses. We measure $H$ (harmonic mean) on 4 datasets. Different colors of curves represent different datasets.}
\label{fig:alphabeta-curve}
\end{figure*}

\textbf{Inference about weighting coefficient $\alpha$, $\beta$ and $\lambda$.} These three hyper-parameters are the weighting coefficient of our loss. In this experiment, we vary $\alpha$ from 1 to 5, $\beta$ from 1 to 12 for all datasets, and $\lambda$ from 285 to 305 for CUB, SUN, AwA1, from 70 to 90 for AwA2, the detailed results are shown in Figure \ref{fig:alphabeta-curve}. From $\alpha$-curve of all datasets, we can see that $H$ generally keeps rising along with the increase of $\alpha$. From $\beta$-curve of all datasets, we can see that $H$ rises smoothly on CUB, SUN, and fluctuates greatly on AwA1, AwA2 with the increase of $\beta$. From $\lambda$-curve of all datasets, we can see that performance fluctuates greatly with the increase of classification weighting coefficient $\lambda$, we can conclude that extracting classification subspace features is more helpful to GZSC than extracting common space features.

\begin{figure}[t]
\centering
\includegraphics[scale=0.49]{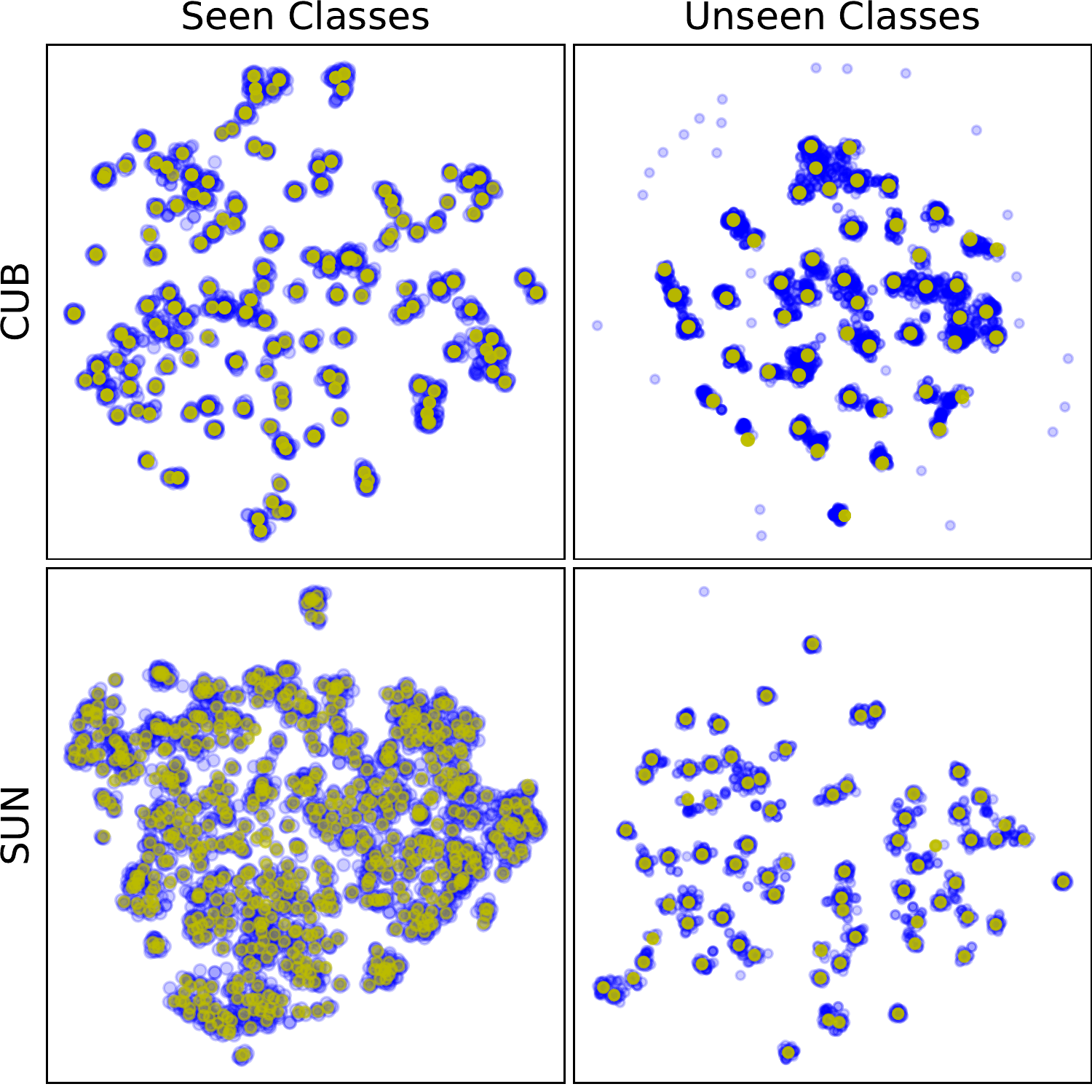}
\caption{t-SNE visualization of cross-modal features alignment on CUB. The visual features in blue, and the semantic features in yellow. (Left: seen classes. Right: unseen classes)}
\label{fig:visualization}
\end{figure}

\textbf{Visualization.} According to the method of qualitatively measuring the extent of two domains \cite{DBLP:iccv/SaitoKSDS19}, we leverage t-SNE \cite{GOOGLE:van2008visualizing} to visualize the features distribution of visual modality and semantic modality in the latent space on CUB and SUN, to visually show that ACMR our proposed can learn well aligned cross-modal representation, as shown in Figure \ref{fig:visualization}. We extract the latent features of seen and unseen classes on testing datasets by the trained ACMR respectively, and use blue dots to represent the features of the visual modality, and yellow dots to represent the features of the semantic modality. From Seen Classes column of CUB and SUN, we can observe that visual features and semantic features overlap almost completely with each other. From Unseen Classes column, although some semantic features have slightly shifted, we can observe that features of two modalities also coincide well. Hence, it verifies the effectiveness of our proposed ACMR.

\section{Conclusion}
In this work, we proposed an innovative autoencoder network via learning aligned cross-modal representations (ACMR) for generalized zero-shot classification (GZSC). Our main contributions lie in innovative Vision-Semantic Alignment (VSA) and Information Enhancement Module (IEM). VSA strengthens the alignment of cross-modal latent features on latent subspace by a learned classifier, while IEM reduces the possibility of latent variables collapse and improve the discriminative ability of latent variables in respective modalities. Extensive experiments on four publicly available datasets demonstrate the state-of-the-art performance of our method and verify the effectiveness of our aligned cross-modal representation on latent subspace guided by classification.

\section{Acknowledgements} 
This research was supported by the National Key Research and Development Program of China (2020AAA09701), National Science Fund for Distinguished Young Scholars (62125601), National Natural Science Foundation of China (62172035, 62006018, 61806017, 61976098), and Fundamental Research Funds for the Central Universities (FRF-NP-20-02).

\bibliography{references}

\end{document}